\documentclass[runningheads]{llncs}
\usepackage{graphicx}
\usepackage{amsmath}
\usepackage{amssymb}
\usepackage{url}
\usepackage{threeparttable}
\usepackage{framed,multirow}
\usepackage{multirow}
\usepackage{gensymb}
\usepackage{hhline}
\usepackage{rotating,graphicx}
\usepackage{ulem}
\usepackage{babel}
\usepackage{amsmath}
\usepackage{mathtools}
\usepackage{graphicx}
\usepackage{hyperref}
\usepackage[title]{appendix}
\usepackage{tabularx, booktabs}
\newtheorem{prop}{Observation}
\begin{document}
\sloppy
\title{Training $\beta$-VAE by Aggregating a Learned Gaussian Posterior with a Decoupled Decoder \thanks{This work was supported by the REACT-EU project KITE (Plattform für KI-Translation Essen).}}
\titlerunning{Training $\beta$-VAE}
%
\author{Jianning Li\inst{1,2} \and
Jana Fragemann\inst{1} \and
Seyed-Ahmad Ahmadi\inst{3} \and
Jens Kleesiek\inst{1} \and
Jan Egger\inst{1,2}}

\authorrunning{Li, J. et al.}
%
\institute{Institute for AI in Medicine, University Medicine Essen, Essen, Germany \email{Jianning.Li@uk-essen.de}\\ \and
Institute of Computer Graphics and Vision, Graz University of Technology, Graz, Austria
\and
NVIDIA, Munich, Germany\\}

\maketitle              
\setcounter{footnote}{0} 
\begin{abstract}
The reconstruction loss and the Kullback-Leibler divergence (KLD) loss in a variational autoencoder (VAE) often play antagonistic roles, and tuning the weight of the KLD loss in $\beta$-VAE to achieve a balance between the two losses is a tricky and dataset-specific task. As a result, current practices in VAE training often result in a trade-off between the reconstruction fidelity and the continuity$/$disentanglement of the latent space, if the weight $\beta$ is not carefully tuned. In this paper, we present intuitions and a careful analysis of the antagonistic mechanism of the two losses, and propose, based on the insights, a simple yet effective two-stage method for training a VAE. Specifically, the method aggregates a learned Gaussian posterior $z \sim q_{\theta} (z|x)$ with a decoder decoupled from the KLD loss, which is trained to learn a new conditional distribution $p_{\phi} (x|z)$ of the input data $x$. Experimentally, we show that the aggregated VAE maximally satisfies the Gaussian assumption about the latent space, while still achieves a reconstruction error comparable to when the latent space is only loosely regularized by $\mathcal{N}(\mathbf{0},I)$.  The proposed approach does not require hyperparameter (i.e., the KLD weight $\beta$) tuning given a specific dataset as required in common VAE training practices. We evaluate the method using a medical dataset intended for 3D skull reconstruction and shape completion, and the results indicate promising generative capabilities of the VAE trained using the proposed method. Besides, through guided manipulation of the latent variables, we establish a connection between existing autoencoder (AE)-based approaches and generative approaches, such as VAE, for the shape completion problem. Codes and pre-trained weights are available at \url{https://github.com/Jianningli/skullVAE}.

\keywords{VAE  \and Disentanglement \and Latent Representation  \and Skull Reconstruction \and Shape Completion}
\end{abstract}

\section{Introduction}
\label{introduction}
Researches on autoencoder (AE)-based generative frameworks such as variational autoencoder (VAE) \cite{kingma2013auto},  $\beta$-VAE \cite{higgins2016beta} and their enhancements \cite{razavi2019generating,van2017neural,dai2019diagnosing} have gained tremendous progress over the years, and the gap with Generative Adversarial Nets (GANs) \cite{goodfellow2014generative} with respect to generative quality has been significantly reduced. The theoretical implication of VAE's success is even far-reaching: high generative quality is achievable without resorting to adversarial training as in GANs. However, VAE has been faced with persistent challenges in practice $-$ given a randomly complex dataset, it is tricky and non-trivial to properly tune the weights between the reconstruction loss, which is responsible for high-quality reconstruction, and the Kullback-Leibler divergence (KLD) loss, which theoretically guarantees a Gaussian and continuous latent space \cite{burgess2018understanding}. High-performing generative models shall satisfy both requirements \cite{makhzani2015adversarial}. It's nevertheless popularly believed that enforcing the VAE's Gaussian assumption about the latent space by applying a large weight $\beta$ on the KLD loss tends to compromise the reconstruction fidelity, leading to a trade-off between the two, as demonstrated in both theory \cite{kim2018disentangling,asperti2020balancing} and practice \cite{bowman2015generating}. Efforts in solving the problem go primarily in three directions: (a) Increase the complexity of the latent space by using a Gaussian mixture model (GMM) \cite{dilokthanakul2016deep,guo2020variational}, or loosen the constraint on the identity covariance matrix \cite{langley2022structured}. These measures aim to increase the capacity and flexibility of the latent space to allow more complex latent distributions, thus reducing the potential conflicts between the dataset and the prior latent assumption;  (b) From an information theory perspective, increase the mutual information between the data and the latent variables explicitly \cite{rezaabad2020learning,hoffman2016elbo,serdega2020vmi,qian2019enhancing,chen2016infogan} or implicitly \cite{zhao2019infovae,dieng2019avoiding}. For example, \cite{dieng2019avoiding} uses skip connections between the output data space (i.e., the ground truth) and the latent space and proves, both analytically and experimentally, that such skip connections increases the mutual information. These methods are motivated by the observation that the Gaussian assumption about the latent space might prevents the information about the data being efficiently transmitted from the data space to the latent space, and thus the resulting latent variables are not sufficiently informative for a subsequent authentic reconstruction; (c) Most intuitively, tune the weight $\beta$ of the KLD loss manually (i.e., trial and error) or automatically \cite{sandfort2021use,bowman2015generating,dai2019diagnosing} during optimization in order to achieve a balance between the reconstruction quality and the latent Gaussian assumption. Specifically,  \cite{sandfort2021use} and \cite{bowman2015generating} applies an increasing weight on the KLD loss during training. A small $\beta$ at the early stage of training allows the reconstruction loss to prevail, so that the latent variables are informative about the reconstruction. A large $\beta$ at a later stage attends to the Gaussian assumption. \cite{dai2019diagnosing} goes a step further and proposes to learn the balancing weight directly. Conceptually, the method presented in our paper is born out of a mixture of (a), (b) and (c), in that our method acknowledges the opposite effects the reconstruction loss and KLD loss may have while training a $\beta$-VAE (a,b), and we try to solve the problem by modulating the weight $\beta$ (c).  Among existing methods, it is worth mentioning that the two-stage method proposed in \cite{dai2019diagnosing} is very similar to our method in form, nevertheless, in essence they differ fundamentally in many aspects \footnote{\cite{dai2019diagnosing} was brought to our attention by a key word search ‘two stage vae training’, after the completion of our method. An empirical comparison between \cite{dai2019diagnosing} and our method is provided.}: (i) In the first stage, \cite{dai2019diagnosing} trained a VAE for small reconstruction errors, without requiring the posterior to be close to $\mathcal{N}(\mathbf{0},I)$, by lowering the dimension of the latent space,  whereas we used a large weight $\beta$ for the KLD loss in this stage to ensure that the posterior maximally approximates $\mathcal{N}(\mathbf{0},I)$, which nevertheless leads to a large reconstruction error (i.e., low reconstruction fidelity), and the latent space dimension was not accounted for. (ii) In the second stage, \cite{dai2019diagnosing} trained another VAE using $\mathcal{N}(\mathbf{0},I)$ as the ground truth latent distribution, and the VAE from both steps are aggregated to achieve a small reconstruction error and a latent distribution close to $\mathcal{N}(\mathbf{0},I)$ in a unified model. In our method, however, we trained a decoupled decoder with independent parameters using the Gaussian variables from the first stage as input and the data as the ground truth. The decoder is able to converge to a small reconstruction error. \cite{dai2019diagnosing} and our method are similar in that we try to meet the two criteria of a high-performing generative model i.e., small reconstruction errors and continuous Gaussian latent space $\mathcal{N}(\mathbf{0},I)$ in two separate stages. However, besides the obvious difference that we realize the two goals in reversed order, the technical and theoretical implications also differ, as will be presented in detail in the following sections.

To put theories into practice, we evaluated the proposed method on a medical dataset $-$ SkullFix \cite{kodym2021skullbreak}, which is curated to support researches on 3D human skull reconstruction and completion \cite{li2021autoimplant,li2021automatic,Li2022}. Unlike commonly used VAE benchmarks, such as CIFAR-10, CelebA and MNIST, which are non-medical, relatively lightweight and in 2D, the medical images we used are of high resolution and in 3D. Besides the intended use of VAE for skull reconstruction, the trained VAE is capable of producing a complete skull given as input a defective skull, by slightly modifying the latent variables of the input, similar to a denoising VAE \footnote{A defective skull can be seen as a complete skull injected with \textit{noise}, i.e., a defect.} \cite{im2017denoising}. We can thus establish a connection between conventional AE-based skull shape completion approaches \cite{li2020baseline,li2021autoimplant,li2021automatic,Li2022} and AE-based generative models such as VAE. The contribution and organization of our paper is summarized as follows:
\begin{enumerate}
\item We presented a concise review and a careful analysis of the common \textit{Reconstruction-KLD} balance problems in $\beta$-VAE, and reformulated the problem in a way such that a solution can be intuitively derived. (Section \ref{introduction}, \ref{beta_vae_} and \ref{section_antagonistic_mechanism})
\item We proposed a simple and intuitive two-stage method to train a $\beta$-VAE for maximal generative capability i.e., high-fidelity reconstruction with a continuous and unit Gaussian latent space. 
The proposed method is free from tuning the hyper-parameter ($\beta$) given a specific dataset. (Section \ref{two_stage})
\item We established a connection between existing AE-based shape completion methods and generative approaches. Results are promising in both quantitative and empirical aspects. (Section \ref{application})
\item The proposed method and the results bear empirical implications: an image dataset, even if complex in the image space, can be maximally mapped to a lower dimensional, continuous and unit Gaussian latent space by imposing a large $\beta$ on the KLD loss. The resulting latent Gaussian variables, even if their distributions are highly overlapped in the latent space, can be mapped to the original image space with high variations and fidelity, by training a decoder decoupled from the KLD loss, i.e., a decoder trained using only the reconstruction loss. (Section \ref{application} and Section \ref{discussion})
\end{enumerate}

 \section{$\beta$-VAE}
 \label{beta_vae_}
 In the setting of variational Bayesian inference, the intractable posterior distribution $p(z|x)$ of data $x$ is approximated using a parameterized distribution $q_{\theta}(z|x)$,  by optimizing the Kullback-Leibler divergence (KLD) between the two distributions \cite{kingma2013auto}. Under the setting, $log \, p(x)$ i.e., the log-evidence of the observations $x$, is lower bounded by a reconstruction term $E_{x \sim q_{\theta}(x|z)} \, [log \,p_{\phi}(x|z)]$ and a KLD term $D_{KL}(q_{\theta}(z|x)||p(z))$:
 
\begin{equation}
log \, p (x) \geqslant   E_{x \sim q_{\theta}(z|x)} \, [log \,p_{\phi}(x|z)]-D_{KL}(q_{\theta}(z|x)||p(z))
\label{elbo}
\end{equation}
 
 Therefore, the log-evidence $log \, p (x)$ can be maximized by maximizing its lower bound i.e., the right hand side (RHS) of inequality \ref{elbo}, which is commonly known as the evidence lower bound (ELBO). The reconstruction term measures the fidelity of the reconstructed data given latent variables $z$, and the KLD term regularizes the estimated posterior distribution to be $p(z)$, which is generally assumed to be a standard Gaussian i.e., $z \in \mathcal{N}(\mathbf{0},I)$ in VAE. $q_{\theta}(z|x)$ and $p_{\phi} (x|z)$ represent the VAE's encoder and decoder, respectively.  
 
 $\beta$-VAE \cite{higgins2016beta} imposes a weight $\beta$ on the KLD term. A large $\beta$ penalizes the KLD and ensures that the posterior $q_{\theta}(z|x)$ maximally approximates the prior $p(z)=\mathcal{N}(\mathbf{0},I)$ and has a diagonal covariance matrix. The diagonality of the covariance matrix implies that the dimensions representing the latent features of the input data are uncorrelated, which facilitates deliberate and guided manipulation of the latent variables towards a desired reconstruction.
 
\section{The Antagonistic Mechanism of the Reconstruction loss and KLD loss in $\beta$-VAE}
\label{section_antagonistic_mechanism} 

It is well established, experimentally and potentially theoretically, that $\beta$-VAE improves the disentanglement of the latent space desirable for generative tasks by increasing the weight $\beta$ of the KLD term, which however undermines the reconstructive capability \cite{kim2018disentangling}. Here, we interpret the phenomenon by assuming that the reconstruction and KLD term in ELBO are antagonistic and often cannot be jointly optimized for certain hand-curated datasets $-$ the Gaussian assumption about the latent space defies the \textit{inherent} distribution of the dataset, and optimizing KLD exacerbates the reconstruction problem and vice versa. The problem could be seen more often when it comes to complex and high-dimensional 3D medical images in the medical domain \cite{fragemann2022review}. In this section, we give intuitions of the antagonistic mechanism of the reconstruction and KLD term, from the perspective of information theory and machine learning.
 
\subsection{Information Theory Perspective}
The interpretation of VAE's objective (ELBO) can be closely linked with information theory \cite{alemi2016deep,chechik2003information,rezaabad2020learning,burgess2018understanding}.  In \cite{burgess2018understanding,alemi2016deep}, the authors present intuitions for the connection between $\beta$-VAE's objective and the information bottleneck principle \cite{alemi2016deep,chechik2003information}, and consider the posterior distribution $q_{\theta}(z|x)$ to be a reconstruction bottleneck. Intuitively, since a VAE is trained to reconstruct exactly its input data thanks to the ELBO's reconstruction term, we can think of the unsupervised training as an information transmission process $-$ forcing the posterior $q_{\theta}(z|x)$ to match a standard Gaussian $\mathcal{N}(\mathbf{0},I)$ by using a large $\beta$ causes loss of information about the data in the encoding phase, and as a result, the latent variables $z$ do not carry sufficient information for an authentic reconstruction of the data in the subsequent decoding phase. From an information theory perspective, the KLD between the estimated distribution $q_{\theta}(z|x)$ and the assumed distribution $p(z)$ can be expressed as the difference between their cross-entropy $H(q_{\theta}(z|x)||p(z))$ and the entropy of the posterior $H(q_{\theta}(z|x))$:
 
 \begin{equation}
 \begin{split}
 D_{KL}(q_{\theta}(z|x)||p(z)) & =H(q_{\theta}(z|x), p(z)) - H(q_{\theta}(z|x)) \\
 &=-\int q_{\theta}(z|x) \, log \, p(z) \,dz \,-\,(-\int q_{\theta}(z|x)\,log\,q_{\theta}(z|x) \, dz ) \\
 & = -\int q_{\theta}(z|x)\, log \,\frac{p(z)}{q_{\theta}(z|x)}\,dz
 \label{kld_eq}
 \end{split}
 \end{equation}

Equation \ref{kld_eq} is known as the reverse KLD in VAE optimization. Analogously, the forward KLD is:
 \begin{equation}
 \begin{split}
D_{KL}(p(z)||q_{\theta}(z|x)) & =H(p(z), q_{\theta}(z|x)) - H(p(z)) \\
& =-\int p(z) log\, \frac{q_{\theta}(z|x)}{p(z)} \\
 \label{forward_kld}
 \end{split}
 \end{equation}

Note that KLD is not a symmetric measure i.e., $D_{KL}(p(z)||q_{\theta}(z|x))\neq D_{KL}(q_{\theta}(z|x)||p(z))$. Since $p(z)=\mathcal{N}(\mathbf{0},I)$ is fixed, the forward KLD and cross-entropy are essentially differing only by an additive constant $H(p(z))$. By this, we can connect the optimization of the KLD with the information transmission analogy: the cross-entropy $H(p(z), q_{\theta}(z|x))$ can be interpreted as the (average) extra amount of information needed to transmit, in order to transmit $x$ from the data space to the latent space $z$, using the estimated distribution $q_{\theta}(z|x)$ instead of using $p(z)$. Therefore, optimizing the KLD term minimizes the extra informational efforts, by forcing $q_{\theta}(z|x)=p(z)$. 

Next, we consider the informativeness of the latent variables $z$ towards the input data $x$, which is modelled by their \textit{mutual information} $\mathcal{I}(x,z)$. Using the reverse KLD in Equation \ref{kld_eq} as an example, the expectation of the KLD term with respect to the data $x$ is given by \cite{kim2018disentangling}: 

 \begin{equation}
 \begin{split}
E_{x \sim p_{(x)}}[D_{KL}(q_{\theta}(z|x)||p(z))]& = E_{x \sim p_{(x)}}[ q_{\theta}(z|x)\, log \,(\frac{q_{\theta}(z|x)}{p(z)})] \\
&=  E_{x \sim p_{(x)}} E_{\sim q_{\theta}(z|x)}[log \, \frac{q_{\theta}(z|x)}{p(z)}] \\
&=  E_{x \sim p_{(x)}} E_{\sim q_{\theta}(z|x)}[log \, \frac{q_{\theta}(z|x)}{q(z)} + log \, \frac{q(z)}{p(z)}] \\
&= E_{\sim q_{\theta}(x,z)}[log \,\frac{q_{\theta}(z|x)}{q(z)}] +E_{\sim q(z)} [log \, \frac{q(z)}{p(z)} ] \\
&= I_{\sim q_{\theta}(z|x)} (x,z) + D_{KL} (q(z)||p(z))
\label{mutual_info}
 \end{split}
 \end{equation}

$D_{KL} (q(z)||p(z)) \geqslant 0$ so that  $E_{\sim p_{(x)}}[D_{KL}(q_{\theta}(z|x)||p(z))] \geqslant I_{\sim q_{\theta}(z|x)} (x,z)$. Therefore, the expectation of the KLD term is a upper bound of the mutual information between the data and the latent variables. Consequently, minimizing the KLD also minimizes the mutual information, and hence reduces the informativeness of the latent variables $z$ with respect to the data $x$. Using a larger $\beta$ penalizes the KLD and further reduces the mutual information $I(x,z)$ during optimization. In information transmission analogy, a larger $\beta$ prevents the information being transmitted from the data space to the latent space efficiently.

\subsection{Machine Learning Perspective}
\label{machine_learning_perspective}
\begin{figure}[ht]
\centering
\includegraphics[width=0.7\linewidth]{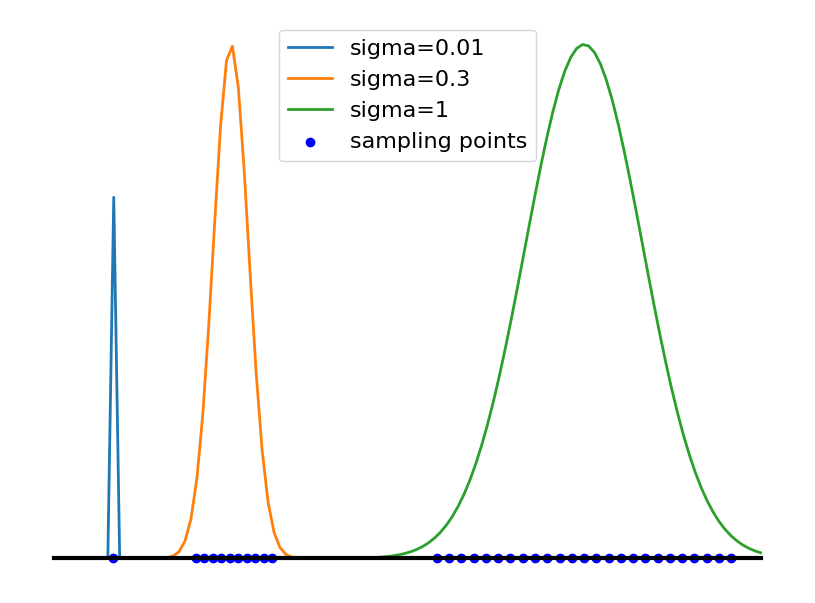}
\caption{1-dimensional Gaussian distributions with different standard deviations $sigma$ ($\sigma_i$).}
\label{fig:sigma}
\end{figure}

The antagonistic mechanism can also be interpreted from a machine learning perspective - a large $\beta$ makes the \textit{latent-to-image} transformation harder to learn. 
Let the posterior Gaussian distribution be $q_{\theta}(z|x) = \mathcal{N}(\mu, \Sigma)$. $\Sigma$ is a diagonal covariance matrix $\Sigma=diag(\sigma_1^2,...,\sigma_d^2)$ and  $\mu=(\mu_1,...,\mu_d)$, $z \in R^d$. In the decoding phase of a VAE, a decoder learns to map $z$ sampled from $\mathcal{N}(\mu, \Sigma)$ \footnote{Reparameterization: $z \in \mathcal{N}(\mu,\Sigma)  => z=\mu+ \sigma \bigodot \varepsilon$, $\varepsilon \in \mathcal{N}(\mathbf{0},I)$.} to the image space. An increase in $\beta$ leads to a larger $\sigma_i$, and hence a broader distribution as shown in Figure \ref{fig:sigma}. When $\begin{Bmatrix}\sigma_i\end{Bmatrix}_{i=1,2,...,d}\rightarrow0$, the distribution shrinks to a single point, and therefore the decoder essentially is tasked with learning a \textit{one-to-one} mapping as in a conventional autoencoder. When $\begin{Bmatrix}\sigma_i\end{Bmatrix}_{i=1,2,...,d}\rightarrow1$, the uncertainty of sampling increases as the distribution broadens, and the decoder learns a \textit{many-to-one} mapping i.e., different $z$ sampled from $\mathcal{N}(\mu, \Sigma)$ correspond to the same output, which is significantly harder than learning a \textit{one-to-one} mapping. Learning the \textit{many-to-one} mapping also makes the latent space continuous. Based on the analysis, we present the following observations:

\begin{prop}
Increasing $\beta$ in $\beta$-VAE  increases $\begin{Bmatrix}\sigma_i\end{Bmatrix}_{i=1,2,...,d}$ and decreases the learnability of the reconstruction problem. Besides, a very high latent dimension $d$ leads to a high sampling uncertainty $\Omega$, which further makes the reconstruction problem harder to learn.
\label{machine_learning}
\end{prop}

\noindent \textit{Explanation:} Given $p(z)=\mathcal{N}(\mathbf{0},I)$, the reverse KLD terms in Equation \ref{kld_eq}, which is used in the implementation of $\beta$-VAE, can be further decomposed as:

 \begin{equation}
 \begin{split}
 D_{KL}(q_{\theta}(z|x)||p(z)) & = \frac{1}{2} [log\frac{|I|}{|\Sigma|}-d+tr\left \{ I^{-1}\Sigma \right \}+(\vec{0}-\mu)^TI^{-1}(\vec{0}-\mu)] \\
                                 & = \frac{1}{2} [-log|\Sigma|-d+tr\left \{ \Sigma \right \}+\mu^T\mu] \\
                                 & = -\frac{1}{2} \sum_i^{d} (1+log \,\sigma_i^2)+\frac{1}{2}\sum_i^{d}\sigma_i^2+\frac{1}{2}\sum_i^{d}\mu_i^2
\label{loss_function}
 \end{split}
 \end{equation}

 In conventional VAE implementations, $\sigma_i$ and $\mu_i$ are connected to the input $x$ via: $(\sigma_1,...,\sigma_{d})^T = \theta^T_1\Phi(x) $ and $\mu^T=\theta^T_2\Phi(x) $. $\Phi(x) \in R^m$ represents the non-linear feature transform of $x$, which is realized through several convolutional layers with non-linear activations in VAE's encoder. $\theta_1$, $\theta_2 \in R^{m \times d}$ represent the weights of two separate linear layers following the convolutional layers. We simplify the explanation by considering that the feature transform $\Phi$ is stochastic \footnote{The feature transform learned by the preceding convolutional layers could mostly be replaced by random feature transform \cite{rahimi2007random}.}, and hence only the two linear layers i.e., $\theta_1$ and $\theta_2$, need to be optimized. Here, we are only interested in $\sigma_i$ and $\theta_1$. The gradient of the KLD objective (weighted by $\beta$) with respect to $\theta^i_1$, which is the $i^{th}$ column vector of the weight matrix $\theta_1$, is therefore:

\begin{equation}
\bigtriangledown \beta D_{KL}(\theta^i_1)=\frac{\partial \beta D_{KL}}{\partial \sigma_i}\frac{\partial \sigma_i}{\partial  \theta^i_1}=\beta \frac{\sigma_i^2-1}{\sigma_i}\Phi(x) 
\label{gradient}
\end{equation}

$\theta^i_1 \in R^m$ and $\sigma_i=\Phi(x)\theta^i_1$. The gradient is a vector-valued function i.e., $\bigtriangledown \beta D_{KL}(\theta^i_1): R^m \rightarrow R^m$. Obviously, the gradient vector has the same dimension as $\theta^i_i$ i.e., $\bigtriangledown \beta D_{KL}(\theta^i_1) \in R^m$.  We then consider a phase of training when the linear layer's output $\sigma_i$ is stabilized to range $(0,1)$ \footnote{Due to the \textit{zero-forcing} effect of the reverse KLD (Equation \ref{kld_eq}), the estimated posterior distribution $q_{\theta}(z|x)$ tends to \textit{squeeze} to match $p(z)=\mathcal{N}(\mathbf{0},I)$. Therefore, it is reasonable to assume that $\sigma_i$ stays in the range of $(0,1)$ when the optimization process is stabilized. This is a further assumption (simplification)  of the explanation.}. To proceed with the explanation, we express the vector notations above using the elements of the vectors, as the following: 

\begin{equation}
\begin{split}
\theta^i_1&=[\theta^i_1(1),\cdots,\theta^i_1(k),\cdots, \theta^i_1(m) ]=\begin{Bmatrix}\theta^i_1(k)\end{Bmatrix}_{k=1,2,...,m}\\
\Phi(x)&=[\Phi_1(x),\cdots,\Phi_k(x), \cdots,\Phi_m(x)]=\begin{Bmatrix}\Phi_k(x)\end{Bmatrix}_{k=1,2,...,m}\\
\bigtriangledown \beta D_{KL}(\theta^i_1)&=\begin{bmatrix}
\bigtriangledown \beta D_{KL}\begin{pmatrix}\theta^i_1(1)\end{pmatrix}\\ 
\cdots \\ 
\bigtriangledown \beta D_{KL}\begin{pmatrix}\theta^i_1(k)\end{pmatrix}\\ 
\cdots \\ 
\bigtriangledown \beta D_{KL}\begin{pmatrix}\theta^i_1(m)\end{pmatrix}
\end{bmatrix}=\beta \frac{\sigma_i^2-1}{\sigma_i}\begin{Bmatrix}\Phi_k(x)\end{Bmatrix}_{k=1,2,...,m}
\end{split}
\end{equation}

$\theta^i_1(k)$ and $\Phi_k(x)$ denote the $k^{th}$ element in $\theta^i_1$ and $\Phi(x)$. Therefore, we can express the gradient with respect to a single element of the weight matrix as:

\begin{equation}
\bigtriangledown \beta D_{KL}\begin{pmatrix}\theta^i_1(k)\end{pmatrix}=\beta \frac{\sigma_i^2-1}{\sigma_i}\Phi_k(x)
\label{gradient_single_weight}
\end{equation}

$\bigtriangledown \beta D_{KL}(\theta^i_1(k))$ is a scalar. The summation notation for the vector product $\sigma_i=\Phi(x)\theta^i_1$ is:

\begin{equation}
\begin{split}
    \sigma_i&=\Phi_1(x)\theta^i_1(1)+\cdots +\Phi_k(x)\theta^i_1(k)+\cdots+\Phi_m(x)\theta^i_1(m)\\
    &=\sum_{k=1}^{m}\Phi_k(x)\theta^i_1(k) \in (0,1)
\end{split}
\label{sigma_element}
\end{equation}

Next, we consider a gradient descent (GD) based optimizer, which updates the elements of the weight matrix according to the following rule:

\begin{equation}
    \theta^i_1(k)_{new}=\theta^i_1(k)_{old}-\alpha\bigtriangledown \beta D_{KL}\begin{pmatrix}\theta^i_1(k)\end{pmatrix}
    \label{gradient_descent}
\end{equation}

$\theta^i_1(k)_{new}$ and $\theta^i_1(k)_{old}$ are the new and old values of the weight. $\alpha> 0$ is the learning rate. If $\Phi_k(x)>0$, then $\bigtriangledown \beta D_{KL}\begin{pmatrix}\theta^i_1(k)\end{pmatrix}<0$. According to the GD optimizer (Equation \ref{gradient_descent}), the weight increases i.e., $\theta^i_1(k)_{new}>\theta^i_1(k)_{old}$. Therefore, $\sigma_i$ increases after this update (Equation \ref{sigma_element}). Likewise, if $\Phi_k(x)<0$, then $\bigtriangledown \beta D_{KL}\begin{pmatrix}\theta^i_1(k)\end{pmatrix}>0$, and the weight decreases  $\theta^i_1(k)_{new}<\theta^i_1(k)_{old}$. In this scenario, $\sigma_i$ still increases (Equation \ref{sigma_element}).

Even though different gradient-based optimizers, such as stochastic gradient descent (SGD), Nesterov’s accelerated gradient method, and Adam \cite{kingma2014adam}, have different rules for updating the weights, generally the above demonstration still applies (the weights are updated in the direction of negative gradients to minimize a loss function). Equation \ref{gradient} and Equation \ref{gradient_single_weight} indicate that given $\beta>1$, the magnitude of the gradient term is larger than when $\beta=1$. Thus, the increase of $\sigma_i$ is larger in each iteration during the optimization process compared to using a smaller $\beta$. Therefore, larger $\beta$ in the KLD objective leads to larger $\sigma_i$ and broader distributions, making the sampling uncertainty of each latent dimension greater. It is also intuitive to understand that the overall uncertainty $\Omega$ is positively related to not only the width ($\sigma_i$) of the distribution but also to the latent dimension $d$: $\Omega \propto d, f(\sigma_i)$. $f(\sigma_i)>1$ is an empirical expression representing the number of points that can be sampled from $\mathcal{N}(\mu_i,\sigma_i)$ during training, and $d$ represents the number of such distributions to sample points from. Hence, the wider the distribution and the larger the latent dimension $d$, the greater the sampling uncertainty and the lower the learnability of the reconstruction problem. However, in practice, one should be aware of a trade-off between sampling uncertainty and the number of latent dimensions needed to store the necessary amount of information of the data for a reasonable reconstruction, especially when the dimension of the data $x$ is high.

\section{Aggregate a Learned Gaussian Posterior with a Decoupled Decoder}
\label{two_stage}

One of the important intuitions we derive from the analysis of the antagonistic mechanism in Section \ref{section_antagonistic_mechanism} is that we can enforce a continuous and unit Gaussian latent space by using a large $\beta$. Conversely, we can maximally release the reconstructive capability of a VAE by lifting the Gaussian constraint on the latent space (i.e., by setting $\beta=0$). These intuitions naturally lead to a two-stage training method, which meets the two requirements in two separate steps. A model satisfying both criteria can be aggregated using the separate results:

\begin{equation}
 z\sim q_{\theta}(z|x), x\sim p_{\phi}(x|z)
\label{two_stage_vae}
\end{equation}

Unlike in VAE where the encoder $q_{\theta}(z|x)$ and decoder $p_{\phi}(x|z)$ is optimized jointly, we first learn a posterior distribution $q_{\theta}(z|x)$ that maximally approximates $\mathcal{N}(0,I)$, and then we sample the approximate Gaussian variables from the learned distributions $ z\sim q_{\theta}(z|x)$, and finally we use a decoder decoupled from the VAE framework (and hence the KLD constraint) to learn a distribution of the data conditioned on the sampled Gaussian variables. The reconstruction can thus be $x\sim p_{\phi}(x|z)$. 

\subsection*{Using a VAE to Learn $q_{\theta}(z|x)$}
A posterior $q_{\theta}(z|x)$ that satisfies the Gaussian assumption is easily learned by imposing a large $\beta$ on the KLD loss. Aside from this, it is safe to assume that since the KLD term is optimized simultaneously with a reconstruction term under a VAE framework, the reconstruction term is likely to have an influence on the latent space\slash variables as well. In other words, latent variables $z$ sampled from $q_{\theta}(z|x)$ likely carry some information about the image space \footnote{Here, ``\textit{carry some information about the image space}\,'' is only an empirical statement, and is used to contrast against \textit{completely random Gaussian variables} in GANs.}.

\subsection*{Using a Decoder to Learn $p_{\phi}(x|z)$}
Empirically, we can provide a convergence guarantee for the decoder by revisiting the fundamental implications of GANs:  a vanilla decoder (i.e., the GANs' generator) is capable of learning a transformation between \textit{completely random Gaussian variables} and authentic (natural) images when combined with a discriminator via adversarial training \cite{goodfellow2014generative}. We leverage this insight and present the following empirical conjecture:

\begin{conjecture}
A vanilla decoder can learn such \textit{Gaussian-to-image} transformation through conventional natural training, if the Gaussian variables already carry some information about the images, i.e., the Gaussian variables are not complete random with respect to the image space. 
\label{training}
\end{conjecture}

\noindent This conjecture can be understood empirically. We require that the input Gaussian variables of a generative model are \textit{aware} of the image space instead of being complete random as in GANs, and expect that such variables can be mapped to the image space with adequate quality using only natural training, without resorting to GANs' adversarial training. \\

\noindent We already showed previously that the latent variables $ z\sim q_{\theta}(z|x)$ satisfy the Gaussian assumption and are $aware$ of the image space, so that we can apply Conjecture \ref{training} in our method. A decoder decoupled from the VAE framework can be used to learn $p_{\phi}(x|z)$, by optimizing only the reconstruction term. To sum up, the first stage of training aims for a continuous and Gaussian latent space, and the second stage aims for an optimal reconstruction.

\section{Application to Skull Reconstruction and Shape Completion}
\label{application}
\begin{figure}
\centering
\includegraphics[width=\linewidth]{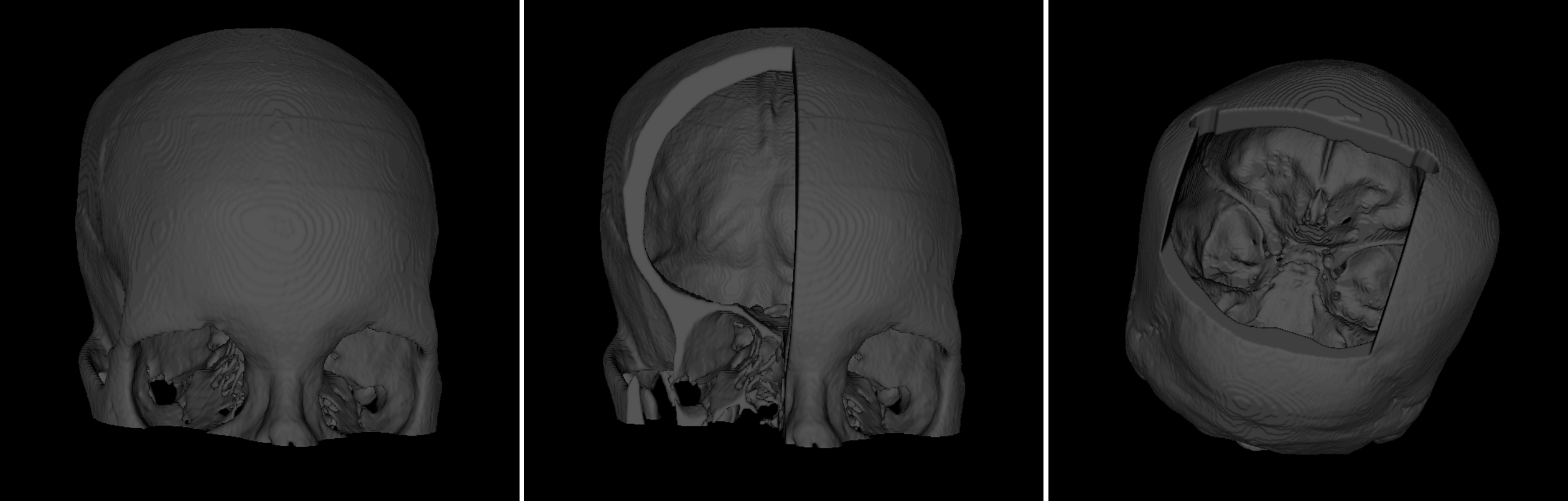}
\caption{Illustration of a complete skull (left) and skulls with facial (middle) and cranial (right) defects.}
\label{fig:dataset}
\end{figure}

In this section, we evaluate the proposed two-stage VAE training method on the SkullFix dataset \cite{kodym2021skullbreak}. The dataset originally contains 100 complete skulls and their corresponding cranially defective skulls for training. For our VAE experiments, we additionally created 100 skulls with facial defects out of the complete skulls, amounting to 300 training samples, and downsampled the images from the original $512\times 512\times Z$ ($Z$ is the axial dimension of the skull images) to $256 \times 256 \times 128$. Figure \ref{fig:dataset} shows a complete skull and its two corresponding defective skulls. We train a VAE on the 300 training samples \footnote{Each sample is used both as input and ground truth as in an autoencoder.}, following the two-stage protocol:
\begin{enumerate}
\item  The VAE is trained for 200 epochs with $\beta=100$.
\item  We fix the encoding and sampling part of the trained VAE from the previous step and train a separate decoupled decoder for \textit{latent-to-image} transformation, for 1200 epochs. 
\end{enumerate}
The VAE's encoder and decoder in Step 1 are composed of six convolutional and deconvolutional layers, respectively. The latent dimension $d$ is set to $32$. The latent vectors $\mu \in R^{32}$ and $(\sigma_1^2,...,\sigma_{32}^2)$ come from two linear layers connecting the encoder's output. The decoder in Step 2 is a conventional 6-layer deconvolutional network that upsamples the 32-dimensional latent variables from Step 1 to the image space ($R^{32} \rightarrow R^{256\times256\times128}$).  For comparison,  we also train the VAE in Step 1 with $\beta=0.0001$ for 200 epochs. In all experiments, Dice loss \cite{li2021automatic} is chosen for the ELBO's reconstruction term. The VAE and related methods are implemented using the MONAI (Medical Open Network for Artificial Intelligence) framework (\url{https://monai.io/}).

\label{skull_rec_training}
 \begin{figure}[ht]
\centering
\includegraphics[width=\linewidth]{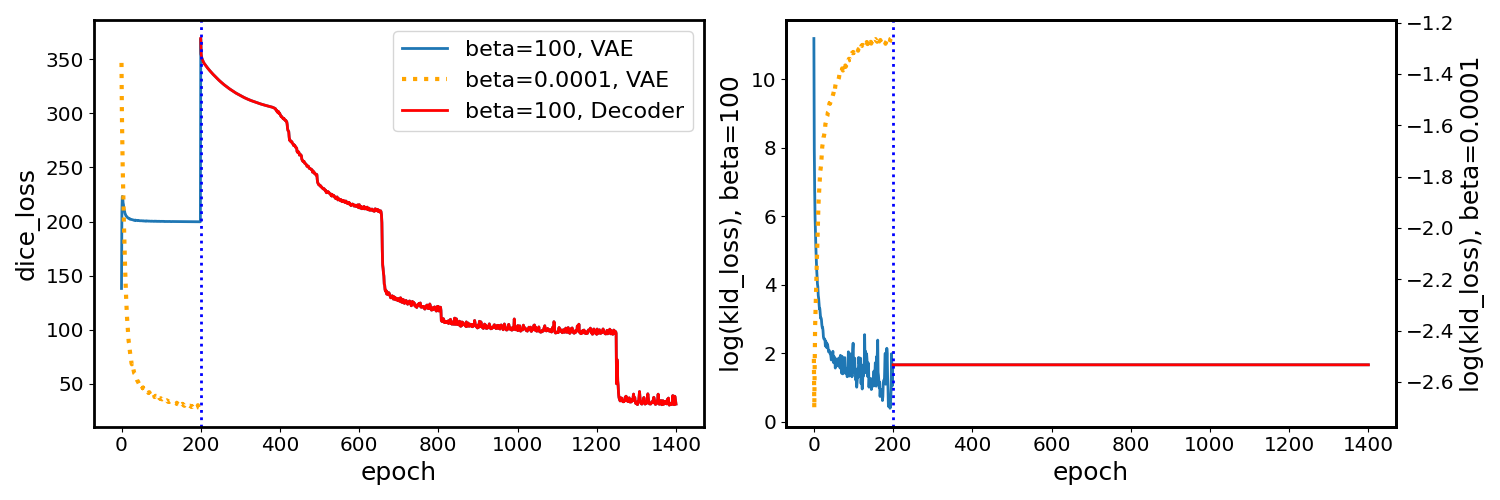}
\caption{Training curves of the VAE and the decoupled decoder regarding Dice (left) and KLD (right) loss. For the KLD loss curve, two y-axis with different y-limits are used for $\beta=100$ and $\beta=0.0001$.}
\label{fig:vae_loss_plots}
\end{figure}

\subsection{Training Curves}

Figure \ref{fig:vae_loss_plots} shows the training curves of the Dice and KLD loss of the regular VAE and decoupled decoder. The curves conform to our analysis and intuitions:

\begin{enumerate}
\item  With $\beta=0.0001$, the reconstruction loss prevails and can decrease to a desired small value, while the KLD loss increases (i.e., antagonistic), which guarantees a high reconstruction fidelity by largely $\textit{ignoring}$ the latent Gaussian assumption.
\item  With $\beta=100$, the KLD loss decreases while the reconstruction loss increases. The VAE guarantees a continuous Gaussian latent space while failing in authentic reconstruction \footnote{We observe that the VAE always gives unvaried output under $\beta=100$, and the output preserves the general skull shape but lacks anatomical (e.g., facial) details. See Figure \ref{fig:rec_cmp} (B) in Appendix \ref{training_curve_1200_epoch} for a reconstruction result under $\beta=100$. This is due to the fact that large $\beta$ broadens the distributions of the latent variables, making different distributions maximally overlap. This tricks the decoder to believe that all latent variables originate from the same distribution, hence giving unvaried reconstruction \cite{burgess2018understanding}.}. Besides, using $\beta=100$ keeps the reconstruction loss from further decreasing (i.e., antagonistic. See Figure \ref{fig:vae_loss_1200_epoch} in Appendix \ref{training_curve_1200_epoch}.).
\item The latent Gaussian variables generated with $\beta=100$ can be mapped to the original image space at similar reconstruction loss as with $\beta=0.0001$, by using an independently trained decoder, implying that even if the distribution of these Gaussian variables might be heavily overlapping in the latent space, the decoder is still able to produce varied reconstructions if properly\slash sufficiently trained (e.g., train for 1200 epochs). The results conform to our Conjecture \ref{training}.
\item We can obtain a generative model that maximally satisfies the VAE's Gaussian assumption about the latent space while still preserving decent reconstruction fidelity by aggregating the encoder of the VAE trained with $\beta=100$ and an independent decoder trained for \textit{latent-to-image} transformation.
\end{enumerate}

It is important to note that, for different training epochs in Step 2, the input of the decoder corresponding to a skull sample can vary due to stochastic sampling $z \in \mathcal{N}(\mu,\Sigma)$, just like in a complete VAE, to ensure that every variable from the continuous latent space can be mapped to a reasonable reconstruction (i.e., \textit{many-to-one} mapping). One can see that the two-stage VAE training method does not involve tuning the hyper-parameter $\beta$, except that we need to empirically choose a large value e.g., $\beta=100$ in our experiments, for Step 1, and is applicable to other datasets out of the box. Besides, the decoder in Step 2 is trained for 1200 epochs until the reconstruction (Dice) loss can converge to a small value comparable to a regular VAE under $\beta=0.0001$, indicating, as discussed in Section \ref{machine_learning_perspective}, that the \textit{latent-to-image} transformation is more difficult to learn. 

\begin{figure}[ht]
\centering
\includegraphics[width=\linewidth]{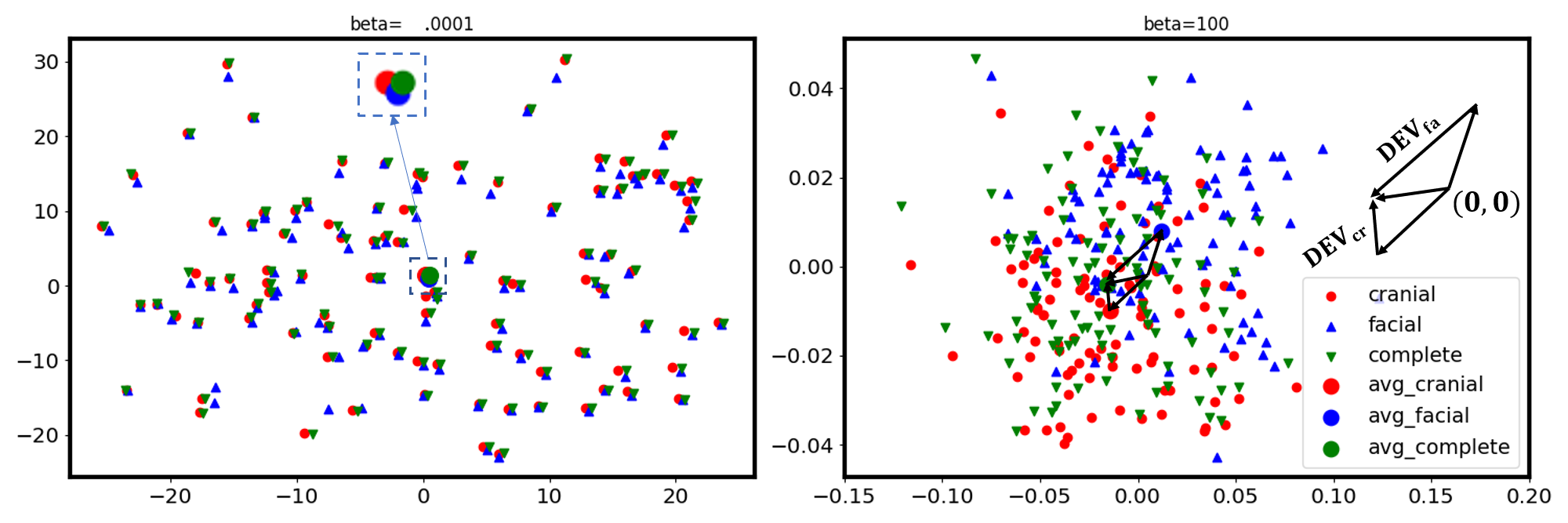}
\caption{The distribution of the latent variables given $\beta=0.0001$ (regular VAE, left) and $\beta=100$ (regular VAE, right). The large filled circles represent the centroids of the respective skull classes. The black arrows on the RHS of the plot point from the origin $(0,0)$ to the centroids, and from the two defective centroids (red and blue) to the complete center (green).}
\label{fig:latent_dist}
\end{figure}

\subsection{Skull Reconstruction and Skull Shape Completion}

\begin{figure}
\centering
\includegraphics[width=\linewidth]{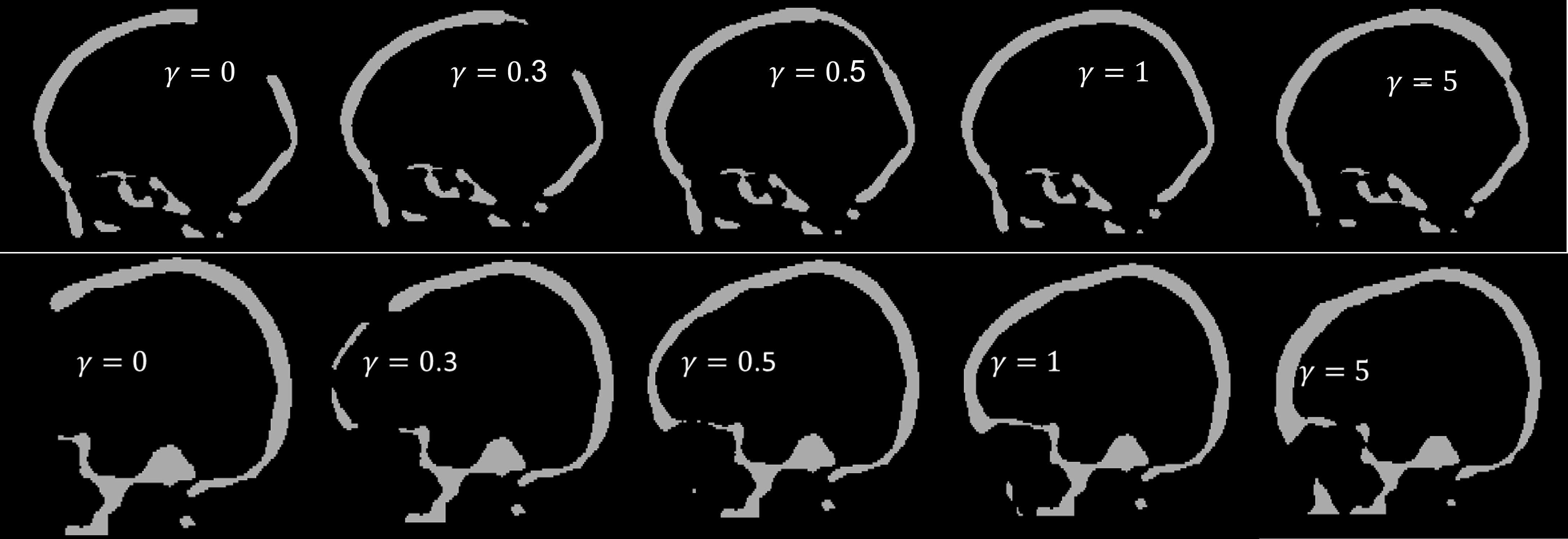}
\caption{Skull shape completion given $\beta=0.0001$ (regular VAE) and different $\gamma$ (see Equation \ref{vector_arithmetic}). The first and second row shows the shape completion results given a cranial and facial defect, respectively.}
\label{fig:small_beta_output}
\end{figure}

In this section, we use the trained regular VAE (under $\beta=0.0001$) and the aggregated VAE (regular VAE under $\beta=100$ aggregated with the trained decoupled decoder) for skull reconstruction and skull shape completion. By skull reconstruction, we evaluate how well the VAE reproduces its input, and by skull shape completion, we use the VAE to generate complete skulls from defective inputs as in \cite{li2021autoimplant},  even if it is not explicitly trained for this purpose. Figure \ref{fig:latent_dist} shows the distribution of the latent variables $z$ of the three skull classes given $\beta=0.0001$ (regular VAE) and $\beta=100$ (regular VAE). For illustration purposes, the dimension of the latent variables is reduced from 32 to 2 using principal component analysis (PCA). We can see that for $\beta=0.0001$, a complete skull and its corresponding two defective skulls are closer to each other than to other skull samples in the latent space, and these variables as a whole are not clustered based on skull classes. For $\beta=100$, we can see that the variables as a whole are packed around the origin $(0,0)$ of the latent space, and are clustered based on different skull classes $-$ complete skulls, skulls with cranial and facial defects. The \textit{cluster-forming} phenomenon indicates that the ELBO's reconstruction term enforces some information related to the image space on the latent variables (Conjecture \ref{training}). The results also generally conform to the VAE's Gaussian assumption $z \in \mathcal{N}(\mathbf{0},I)$, and sampling randomly from the continuous latent space has a higher probability for a reasonable reconstruction than from a nearly discrete latent space ($\beta=0.0001$).  However, unlike the MNIST dataset commonly used in VAE-related researches, the skull clusters are heavily overlapping, with the cluster of the facial defects moderately deviating from the other two. This is also understandable, since we cannot guarantee that the skulls are differently distributed in a latent space while handcrafting the defects from the complete skulls in the image space.

\begin{table*}[ht]
\centering
\caption{Quantitative evaluation results (Dice similarity coefficient - DSC) of skull reconstruction (REC) and skull shape completion (CMP) given $\beta=0.0001$ (regular VAE) and $\beta=100$ (aggregated VAE).}
\begin{tabular}[t]{ccccccccc} 
\toprule
 \multirow{2}{*}{$\beta$} & \multicolumn{3}{c}{Reconstruction (REC)} & \multicolumn{3}{c}{Completion (CMP)}\\
 \cmidrule{2-4}  \cmidrule{6-7} 
 & cranial & facial & complete & & cranial  &  facial  \\ 
\hline
 $\beta=0.0001$ (regular VAE)        &0.9153 & 0.9215&0.9199 &&0.9189&0.9072      \\
 $\beta=100$  (aggregated VAE)           &0.9076 &0.9094 &0.9127 &&0.8978&0.7934 \\
\bottomrule
\end{tabular}
\label{table:quantitative_results}
\end{table*}

\begin{figure}[ht]
\centering
\includegraphics[width=\linewidth]{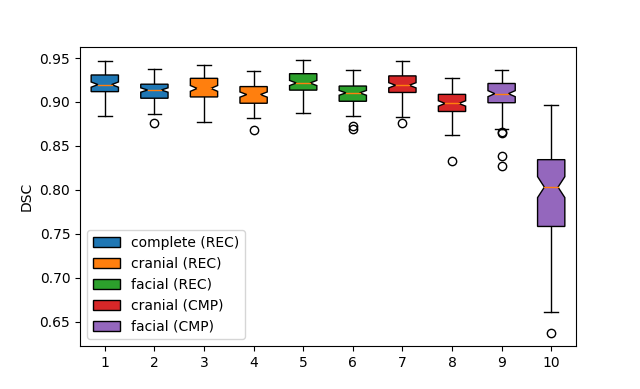}
\caption{DSC boxplots for skull reconstruction (REC) and skull shape completion (CMP) given $\beta=0.0001$ (regular VAE, left) and $\beta=100$ (aggregated VAE, right).}
\label{fig:dsc_boxplot}
\end{figure}

\begin{figure}[ht]
\centering
\includegraphics[width=\linewidth]{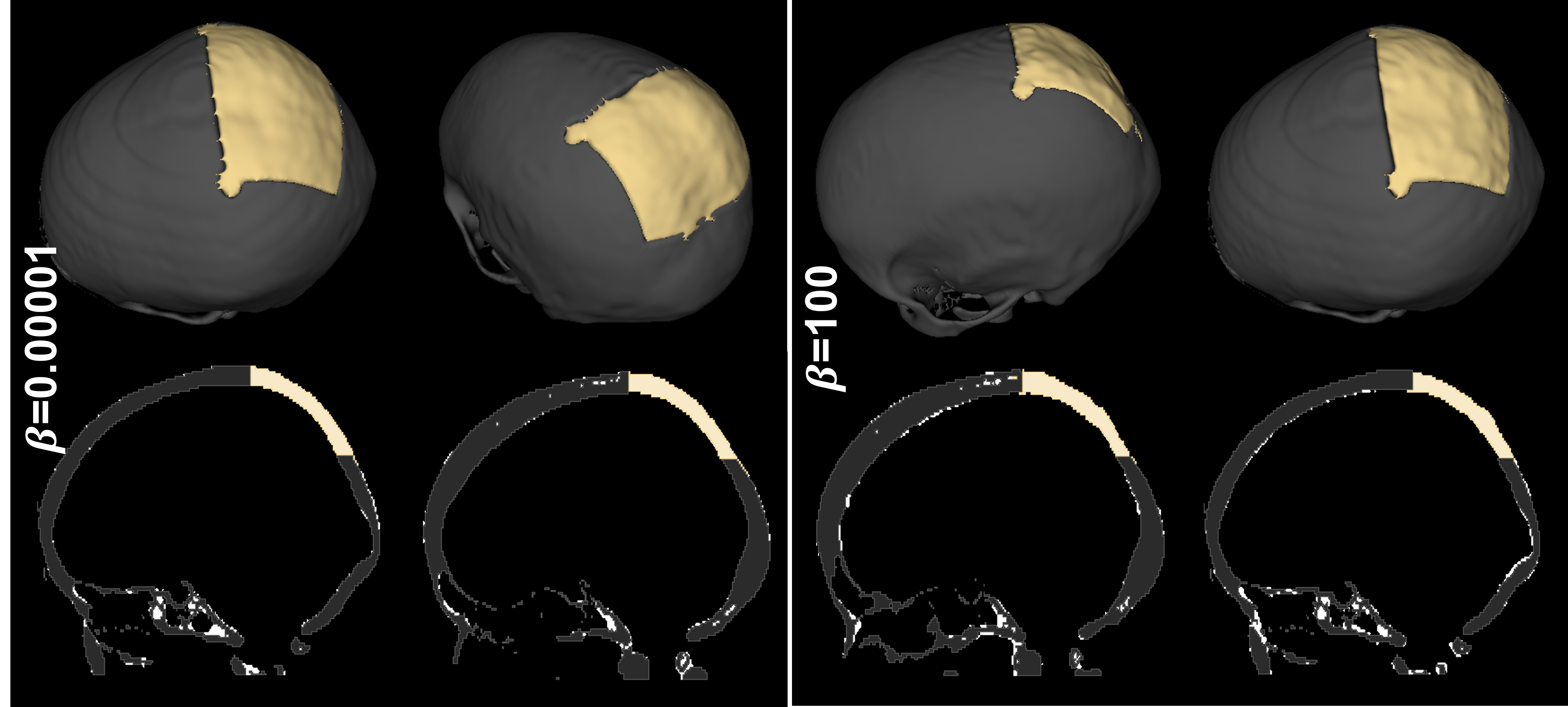}
\caption{Cranial shape completion results given $\beta=0.0001$ (regular VAE) and $\beta=100$ (aggregated VAE). The \textit{implants} shown in yellow correspond to the deviation vectors $DEV_{cr}$ in Equation \ref{deviation}.}
\label{fig:cranial_cmp}
\end{figure}

\begin{figure}[ht]
\centering
\includegraphics[width=\linewidth]{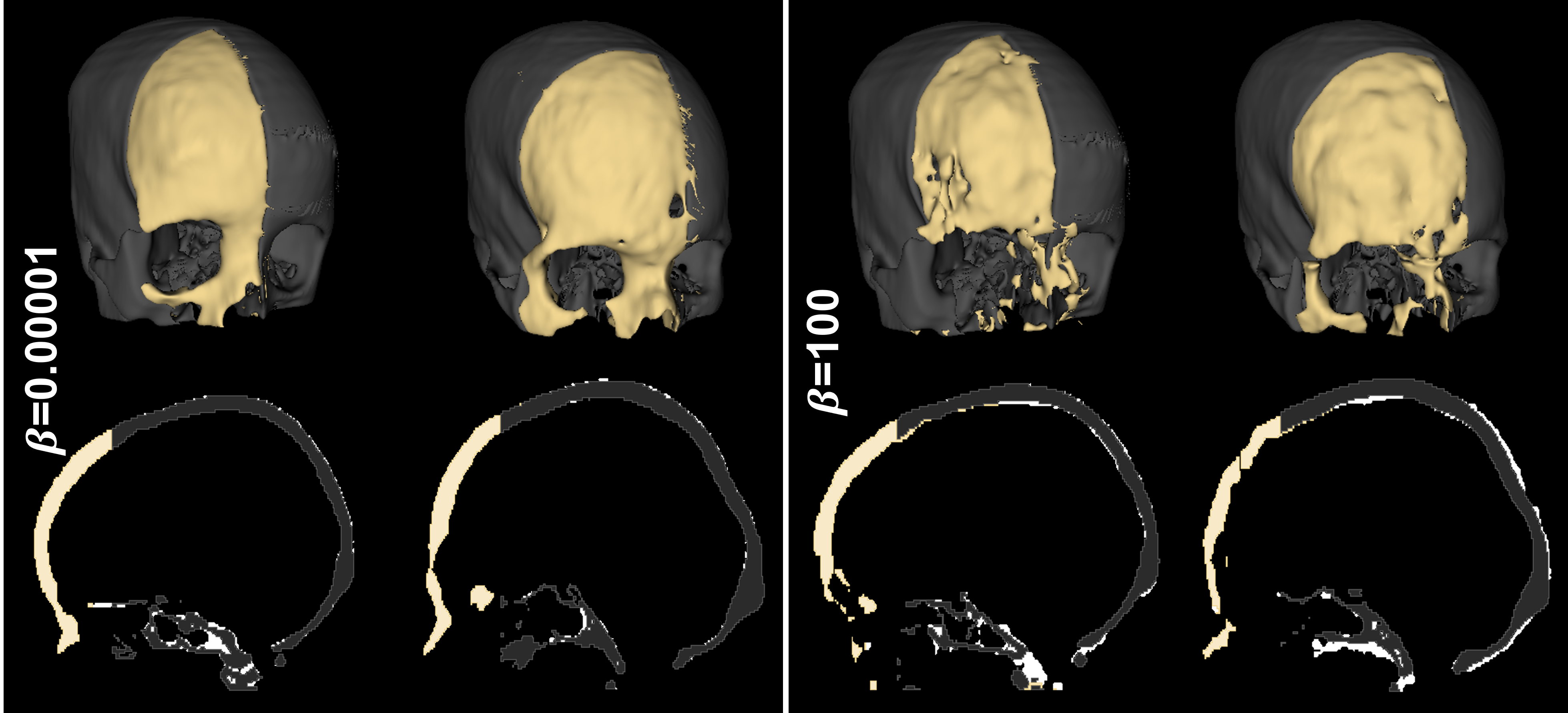}
\caption{Facial shape completion results given $\beta=0.0001$ (regular VAE) and $\beta=100$ (aggregated VAE). The \textit{implants} shown in yellow correspond to the deviation vectors $DEV_{fa}$ in Equation \ref{deviation}.}
\label{fig:facial_cmp}
\end{figure}

For skull reconstruction, these latent variables are mapped to the image space using the respective decoders. For skull shape completion, we define the following: let the latent variables of the complete skulls, facially and cranially defective skulls from the training samples be $\mathbf{z}^{co}$, $\mathbf{z}^{fa}$ and $\mathbf{z}^{cr}$.  We therefore compute the average deviation of the defective skulls (facial: $DEV_{fa}$; cranial: $DEV_{cr}$) from their corresponding complete skulls as:

\begin{equation}
\begin{split}
DEV_{cr}&=\frac{1}{N}\sum_{i}^{N}(\mathbf{z}^{co}_{i} -\mathbf{z}^{cr}_{i}) =\frac{1}{N}\sum_{i}^{N}\mathbf{z}^{co}_{i} -\frac{1}{N}\sum_{i}^{N}\mathbf{z}^{cr}_{i}\\
DEV_{fa}&=\frac{1}{N}\sum_{i}^{N}(\mathbf{z}^{co}_{i} -\mathbf{z}^{fa}_{i}) =\frac{1}{N}\sum_{i}^{N}\mathbf{z}^{co}_{i} -\frac{1}{N}\sum_{i}^{N}\mathbf{z}^{fa}_{i}
\label{deviation}
\end{split}
\end{equation}

$N=100$ is the number of samples of a skull class. Since $\frac{1}{N}\sum_{i}^{N}\mathbf{z}^{co}_{i}$, $\frac{1}{N}\sum_{i}^{N}\mathbf{z}^{cr}_{i}$ and $\frac{1}{N}\sum_{i}^{N}\mathbf{z}^{fa}_{i}$ essentially compute the centroids of the respective skull clusters,  $DEV_{fa}$ and $DEV_{cr}$ in Equation \ref{deviation} can be interpreted intuitively as the two deviation vectors shown in Figure \ref{fig:latent_dist} or the \textit{implant}. In this sense, skull shape completion means interpolating between the defective and complete skull classes. Given a defective skull whose latent variable is $\mathbf{z}_{ts}$ as a test sample, the latent variable that decodes to a complete skull corresponding to the defective test sample can be computed as (vector arithmetic):

\begin{equation}
\begin{split}
z^{cr\rightarrow co}&=\mathbf{z}_{ts}+\gamma DEV_{cr}\\
z^{fa\rightarrow co}&=\mathbf{z}_{ts}+\gamma DEV_{fa}
\label{vector_arithmetic}
\end{split}
\end{equation}

$\gamma \in \mathcal{R}$ controls the extent of completion. With $\gamma=0$, we expect that the resulting latent variable would be decoded to the original defective sample (i.e., skull reconstruction). With $\gamma=1$, we expect that decoding the latent variable yields a complete skull (i.e., skull shape completion). Figure  \ref{fig:small_beta_output} shows the decoding results given different $\gamma$ and $\beta=0.0001$. Given $0<\gamma<1$, we can see the gradual evolution of the output (from defective to complete). Similar results are observed for $\beta=100$ (aggregated VAE). Besides, the VAE can extrapolate the output by \textit{thickening} the \textit{implant} given $\gamma>1$, as can be seen in Figure  \ref{fig:small_beta_output}. Table \ref{table:quantitative_results} and Figure \ref{fig:dsc_boxplot} show the quantitative evaluation results for skull reconstruction and skull shape completion. We measure the agreement between the input and the reconstruction using Dice similarity coefficient (DSC). We can see that the reconstruction accuracy obtained under a heavy regularization of the latent space (i.e., $\beta=100$, aggregated VAE) is comparable to that under a weak regularizer  (i.e., $\beta=0.0001$, regular VAE), indicating that the Gaussian variables can be used for decent skull reconstruction using the decoupled decoder, without resorting to adversarial training as in GANs (Conjecture \ref{training}). However, for $\beta=100$, we see a noticeable performance drop in the shape completion task on facial defects. An educated guess of the cause, based on known results reported in Table \ref{table:quantitative_results} and Figure \ref{fig:dsc_boxplot}, is that in the latent space created by using $\beta=100$, the variables from the complete and cranially defective skulls are heavily overlapped, while the latent variables of the facially defective skulls are obviously deviating (Figure \ref{fig:latent_dist}). As a result, the magnitude of the deviation vector representing the facial \textit{implant} (Equation \ref{deviation}) is much higher compared to that of the cranial deviation vector. Figure \ref{fig:cranial_cmp} and Figure \ref{fig:facial_cmp} show examples of skull shape completion results in 2D and 3D under $\beta=0.0001$ (regular VAE) and $\beta=100$ (aggregated VAE), for the cranial and facial defects. The results are obtained based on Equation \ref{vector_arithmetic} given $\gamma=1$. We further obtain the \textit{implant} by taking the \textit{difference} between the decoding results i.e., completed skulls and the defective input in the image space. For conventional AE-based skull shape completion results, we refer the readers to Figure \ref{fig:facialRec} in Appendix \ref{AE_based_Shape_Completion}.

\section{Discussion and Conclusion}
\label{discussion}
An explicit separation of the prediction accuracy and network complexity in the loss function of variational inference allows one to manipulate the optimization process by adjusting the weights of the two loss terms \cite{graves2011practical}. In VAE, the reconstruction term and the KLD term corresponds to the \textit{accuracy} and \textit{complexity}, respectively. In this paper, we evaluated the influence of the KLD weight $\beta$ on the reconstructive and generative capability of $\beta$-VAE, using a skull dataset. We considered three scenarios:  $\beta=0.0001$ (low network complexity), $\beta=100$ (high network complexity), and an aggregate of the posterior distribution from $\beta=100$ and a decoupled decoder independently trained for skull reconstruction. Experiments reveal that even if the KLD loss increases during training under $\beta=0.0001$, it is able to stabilize at some point (Figure \ref{fig:vae_loss_plots}). As expected, the resulting latent space is non-continuous (Figure \ref{fig:latent_dist}). However, the tendency of the latent variables assembling around the origin $(0,0)$ is noticeable, which shows the (weak) influence of the KLD loss. This could explain the reason why we can still interpolate smoothly among skull classes on a local scale given $\beta=0.0001$, and perform skull shape completion by manipulating the latent variables (Figure \ref{fig:small_beta_output}, Equation \ref{deviation}, Equation \ref{vector_arithmetic}). The reconstructive capability of the network can also be fully exploited by using the small $\beta$ $-$ an advantage (i.e., small reconstruction error) which appears to outweigh the disadvantages (i.e., non-continuous latent space) on this particular dataset and task.  In comparison, we realize a globally continuous, maximally unit Gaussian latent space by using $\beta=100$ (Figure \ref{fig:latent_dist}), and a small reconstruction loss by training a separate decoder decoupled from the KLD loss (Figure \ref{fig:vae_loss_plots}). However, the shape completion performance on facial defects is suboptimal compared to that on cranial defects and skull reconstruction (Table \ref{table:quantitative_results} and Figure \ref{fig:dsc_boxplot}). Empirically, the results could be improved by reformulating the definition of the cluster centroids in Equation \ref{deviation}, so that the centroids of different skull clusters are closer. The problem is not unexpectable, since exploiting the disentanglement features of the latent space is often a delicate issue \cite{fragemann2022review}. Last but not least, we would like to point out one of the major limitations of our current work: the proposed two-stage method in Section \ref{two_stage} is largely suggested by empirical analysis. Rigorous theoretical guarantee has yet been provided in its current form. Future work demonstrating or refuting the universal applicability and theoretical guarantee of the method is to be conducted. Besides, Observation \ref{machine_learning} and Conjecture \ref{training} as well as their explanation presented in this paper are also empirical. A formal expression of the Rademacher complexity involving the latent dimension $d$, the weight $\beta$ and the variance $\sigma_i$ for the learnability of the reconstruction problem (Observation \ref{machine_learning}), as well as a theoretical proof of the convergence guarantee for the decoupled decoder (Conjecture \ref{training}) is also desired in future work.

\bibliographystyle{splncs04}
\bibliography{references}

\newpage

\begin{subappendices}
\renewcommand{\thesection}{\Alph{section}}%

\section{VAE Training Curve (1200 Epochs) under $\beta=100$}
\label{training_curve_1200_epoch}

\begin{figure}
\renewcommand\thefigure{A.1}
\centering
\includegraphics[width=\linewidth]{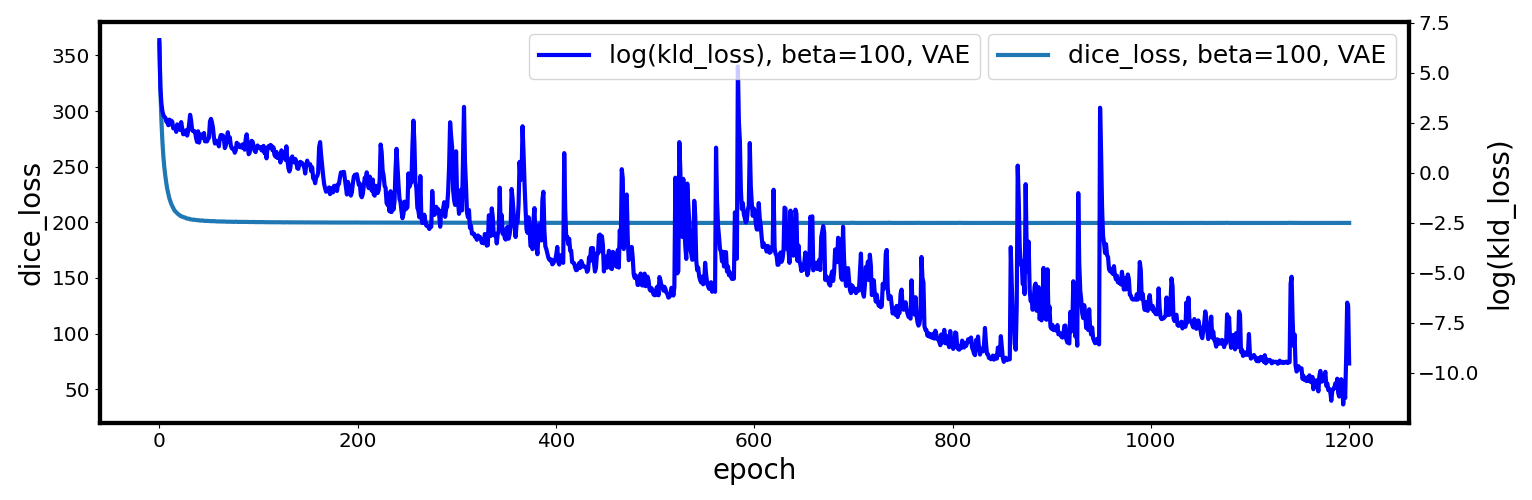}
\caption{Training a regular VAE (the same VAE used in Section \ref{application}) for 1200 epochs under $\beta=100$. The curve shows the Dice and KLD loss for the entire training process.}
\label{fig:vae_loss_1200_epoch}
\end{figure}

\begin{figure}
\renewcommand\thefigure{A.2}
\centering
\includegraphics[width=\linewidth]{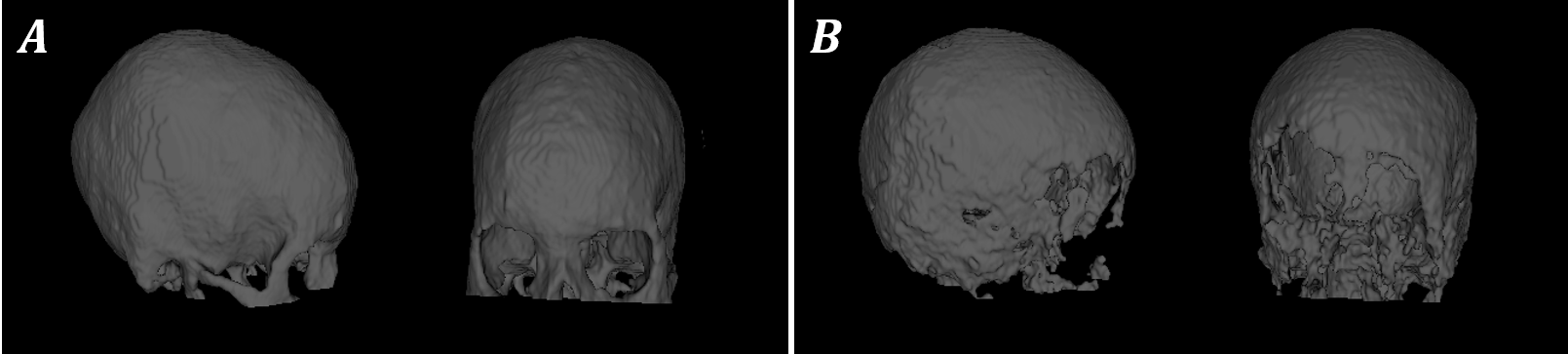}
\caption{Reconstruction results from (A) the aggregated VAE, and (B) a regular VAE trained for 1200 epochs under $\beta=100$.}
\label{fig:rec_cmp}
\end{figure}

Figure \ref{fig:vae_loss_1200_epoch} shows the training curve of the Dice and KLD loss under $\beta=100$. We can see that after 1200 epochs (training took approximately 120 hours on a desktop PC with a 24 GB NVIDIA GeForce RTX 3090 GPU and an AMD Ryzen 9, 5900X 12-Core CPU), the reconstruction (Dice) loss is still not able to decrease to a desirable small value. As a result, the VAE gives poor reconstruction (Figure \ref{fig:rec_cmp} (B)), and the output is unvaried given different input skulls. In contrast, the reconstruction loss of the decoupled decoder can converge to around 30 after 1200 epochs (Figure \ref{fig:vae_loss_plots}), and thus the aggregated VAE can produce desirable and varied reconstructions, as can be seen from Figure \ref{fig:rec_cmp} (A).

\section{AE-based Skull Shape Completion}
\label{AE_based_Shape_Completion}

\begin{figure}[ht]
\renewcommand\thefigure{B.1}
\centering
\includegraphics[width=\linewidth]{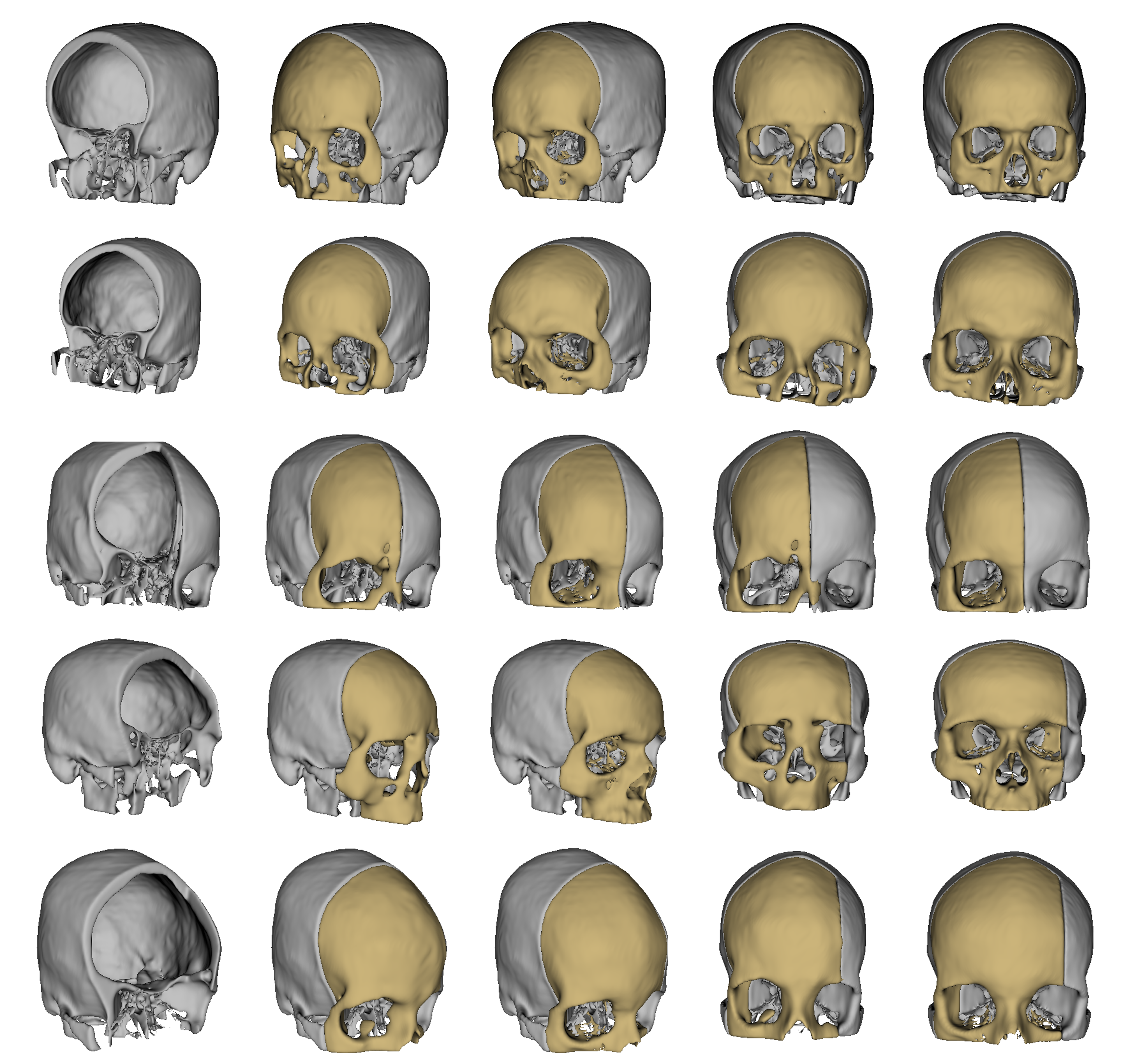}
\caption{Skull shape completion results for facial defects, obtained using a conventional autoencoder from \cite{li2020baseline}. The first column shows the input. The second and fourth column show the predictions from two different views. The third and last column show the corresponding ground truth.}
\label{fig:facialRec}
\end{figure}

In this appendix, we show results of an AE-based shape completion method for facial defects. The AE used here is the same as $N_1$ in \cite{li2020baseline}, and is implemented in tensorflow (\url{https://www.tensorflow.org/}). Unlike VAE, the AE is trained explicitly for skull shape completion, with the input being a facially defective skull and the output being the corresponding complete skull. Qualitative results are shown in Figure \ref{fig:facialRec}. It is obvious and expected that the completion results are much better than those of VAE, since the AE is explicitly trained for the completion task. Furthermore, bear in mind that skull completion (especially for facial defects) is an ill-posed problem, and the network is trained to produce a complete skull that is anatomically plausible but not necessarily resembling the ground truth.

\section{Matrix Notation for $\bigtriangledown \beta D_{KL}(\theta_1)$}
\label{matrix_notation}
The partial derivative of the KLD loss with respect to a single element of the weight matrix $\theta_1$ (i.e, Equation \ref{gradient_single_weight} in the main manuscript) can be more compactly derived and expressed using matrix calculus. To this end, we define:

\begin{equation}
\renewcommand\theequation{C.1}
\theta_1=\begin{bmatrix}
\theta^{11}_1 & \theta^{21}_1  & \theta^{31}_1& \cdots& \theta^{m1}_1\\ 
 \theta^{12}_1 & \theta^{22}_1  & \theta^{32}_1 &\cdots&\theta^{m2}_1\\ 
 \vdots &\vdots   & \vdots &\ddots &\vdots\\ 
 \theta^{1d}_1& \theta^{2d}_1 & \theta^{3d}_1 &\cdots &\theta^{md}_1\\ 
\end{bmatrix}^T \in R^{m\times d}
\label{eq:appendixc1}
\end{equation}

\begin{equation}
\renewcommand\theequation{C.2}
\begin{split}
 \sigma=\theta^T_1 \cdot\Phi(x)& = \begin{bmatrix}
\theta^{11}_1 & \theta^{21}_1  & \theta^{31}_1& \cdots& \theta^{m1}_1\\ 
 \theta^{12}_1 & \theta^{22}_1  & \theta^{32}_1 &\cdots&\theta^{m2}_1\\ 
 \vdots &\vdots   & \vdots &\ddots &\vdots\\ 
 \theta^{1d}_1& \theta^{2d}_1 & \theta^{3d}_1 &\cdots &\theta^{md}_1\\ 
\end{bmatrix} \cdot \begin{bmatrix}\Phi_1(x)\\ \Phi_2(x)\\ \vdots \\ \Phi_m(x)\
\end{bmatrix} \\
& =\begin{bmatrix}
\sum_{k=1}^{m}\theta^{k1}_1\Phi_k(x)\\ 
\sum_{k=1}^{m}\theta^{k2}_1\Phi_k(x)\\ 
\vdots \\ 
\sum_{k=1}^{m}\theta^{kd}_1\Phi_k(x)
\end{bmatrix}=\begin{bmatrix}
\sigma_1\\ 
\sigma_2\\ 
\vdots \\ 
\sigma_d
\end{bmatrix}
\end{split}
\label{eq:appendixc2}
\end{equation}

Based on Equation \ref{loss_function} in the main manuscript and the above matrix notations, we can calculate the derivative of the KLD loss with respect to $\theta_1$ directly:

\begin{equation}
\renewcommand\theequation{C.3}
\begin{split}
\bigtriangledown \beta D_{KL}(\theta_1)&=\beta\begin{bmatrix}
\frac{\partial D_{KL}}{\partial \theta^{11}_1} & \frac{\partial D_{KL}}{\partial \theta^{21}_1}  & \frac{\partial D_{KL}}{\partial \theta^{31}_1}& \cdots& \frac{\partial D_{KL}}{\partial \theta^{m1}_1}\\ 
 \frac{\partial D_{KL}}{\partial \theta^{12}_1} & \frac{\partial D_{KL}}{\partial \theta^{22}_1}  & \frac{\partial D_{KL}}{\partial \theta^{32}_1} &\cdots&\frac{\partial D_{KL}}{\partial \theta^{m2}_1}\\ 
 \vdots &\vdots   & \vdots &\ddots &\vdots\\ 
 \frac{\partial D_{KL}}{\partial \theta^{1d}_1}& \frac{\partial D_{KL}}{\partial \theta^{2d}_1} & \frac{\partial D_{KL}}{\partial \theta^{3d}_1} &\cdots &\frac{\partial D_{KL}}{\partial \theta^{md}_1}\\ 
\end{bmatrix}^T\\
&=\beta\begin{bmatrix}
\frac{\partial D_{KL}}{\partial \sigma_1}\frac{\sigma_1}{\partial \theta^{11}_1} & \frac{\partial D_{KL}}{\partial \sigma_1}\frac{\sigma_1}{\partial \theta^{21}_1}  & \frac{\partial D_{KL}}{\partial \sigma_1}\frac{\sigma_1}{\partial \theta^{31}_1}& \cdots& \frac{\partial D_{KL}}{\partial \sigma_1}\frac{\sigma_1}{\partial \theta^{m1}_1}\\ 
 \frac{\partial D_{KL}}{\partial \sigma_2}\frac{\sigma_2}{\partial \theta^{12}_1} &  \frac{\partial D_{KL}}{\partial \sigma_2}\frac{\sigma_2}{\partial \theta^{22}_1}  & \frac{\partial D_{KL}}{\partial \sigma_2}\frac{\sigma_2}{\partial \theta^{32}_1} &\cdots& \frac{\partial D_{KL}}{\partial \sigma_2}\frac{\sigma_2}{\partial \theta^{m2}_1}\\ 
 \vdots &\vdots   & \vdots &\ddots &\vdots\\ 
 \frac{\partial D_{KL}}{\partial \sigma_d}\frac{\sigma_d}{\partial \theta^{1d}_1} & \frac{\partial D_{KL}}{\partial \sigma_d}\frac{\sigma_d}{\partial \theta^{2d}_1} & \frac{\partial D_{KL}}{\partial \sigma_d}\frac{\sigma_d}{\partial \theta^{3d}_1} &\cdots &\frac{\partial D_{KL}}{\partial \sigma_d}\frac{\sigma_d}{\partial \theta^{md}_1}\\ \end{bmatrix}^T\\
&=\beta\begin{bmatrix}
\frac{\sigma^2_1-1}{\sigma_1}\Phi_1(x) & \frac{\sigma^2_1-1}{\sigma_1}\Phi_2(x) & \frac{\sigma^2_1-1}{\sigma_1}\Phi_3(x) & \cdots  & \frac{\sigma^2_1-1}{\sigma_1}\Phi_m(x)\\ 
\frac{\sigma^2_2-1}{\sigma_2}\Phi_1(x) & \frac{\sigma^2_2-1}{\sigma_2}\Phi_2(x) & \frac{\sigma^2_2-1}{\sigma_2}\Phi_3(x) & \cdots  & \frac{\sigma^2_2-1}{\sigma_2}\Phi_m(x)\\ 
 \vdots &  \vdots &  \vdots & \ddots  &  \vdots\\ 
\frac{\sigma^2_d-1}{\sigma_d}\Phi_1(x) & \frac{\sigma^2_d-1}{\sigma_d}\Phi_2(x) & \frac{\sigma^2_d-1}{\sigma_d}\Phi_3(x) & \cdots  & \frac{\sigma^2_d-1}{\sigma_d}\Phi_m(x)
\end{bmatrix}^T \in R^{m\times d}
\end{split}
\end{equation}

We define $\theta^{ki}_1$ as the weight at the $k^{th}$ row and $i^{th}$ column in the weight matrix $\theta_1$. Then, the partial derivative of the KLD loss with respect to  $\theta^{ki}_1$ is simply $\frac{\partial D_{KL}}{\partial \theta^{ki}_1}=\beta\frac{\sigma^2_i-1}{\sigma_i}\Phi_k(x)$, which is the same as Equation \ref{gradient_single_weight} in the main manuscript.

\end{subappendices}

\end{document}